# HELICOPTER TRACK IDENTIFICATION WITH AUTOENCODER

*Liya Wang, Panta Lucic, Keith Campbell, and*
*Craig Wanke*
*The MITRE Corporation, McLean, VA, 22102, United States*

**Abstract**

Computing power, big data, and advancement of algorithms have led to a renewed interest in artificial intelligence (AI), especially in deep learning (DL). The success of DL largely lies on data representation because different representations can indicate to a degree the different explanatory factors of variation behind the data. In the last few year, the most successful story in DL is supervised learning. However, to apply supervised learning, one challenge is that data labels are expensive to get, noisy, or only partially available. With consideration that we human beings learn in an unsupervised way; self-supervised learning methods have garnered a lot of attention recently. A dominant force in self-supervised learning is the autoencoder, which has multiple uses (e.g., data representation, anomaly detection, denoise). This research explored the application of an autoencoder to learn effective data representation of helicopter flight track data, and then to support helicopter track identification. Our testing results are promising. For example, at Phoenix Deer Valley (DVT) airport, where 70% of recorded flight tracks have missing aircraft types, the autoencoder can help to identify twenty-two times more helicopters than otherwise detectable using rule-based methods; for Grand Canyon West Airport (1G4) airport, the autoencoder can identify thirteen times more helicopters than a current rule-based approach. Our approach can also identify mislabeled aircraft types in the flight track data and find true types for records with pseudo aircraft type labels such as HELO. With improved labelling, studies using these data sets can produce more reliable results.

## I. Introduction

In the past few years, researchers have become increasingly interested in deep learning (DL) due to its success in applications such as computer vision, natural language processing (NLP), and self-driving cars. One of most successful stories in DL is supervised learning, which infers a function from labeled training data consisting of a set of training examples. However, one challenge in supervised learning is that data labels are expensive to get, noisy, or only partially available. With consideration that we humans learn in an unsupervised way, self-supervised learning methods have garnered a lot of attention recently.

In self-supervised learning, the autoencoder approach has been effective. An autoencoder is a neural network designed for unsupervised representation learning. It tries to reconstruct the original input while compressing the data to discover a more efficient representation. The success of DL has largely depended on data representation because different representations can indicate the different explanatory factors of variation behind the data [1]. Besides representation learning, researchers also have developed a variety of autoencoders (e.g., anomaly detection autoencoders, denoising autoencoders, variational autoencoders) to solve various types of problems.

Machine learning (ML) is being adopted in many transportation fields including air transportation. For example, Smith et al. [2] generated a survey of different technology forecasting techniques for complex systems and identified machine learning as useful for providing estimates for future technology predictions; Wang et al. [3] explored logistic regression for an unstable approach risk prediction at a specific distance location to the runway threshold. In addition, a variety of ML algorithms were used to predict aviation demand ([4]); Kim et al. [5] proposed long-short term memory (LSTM) to predict flight delays.

Even with the aforementioned ML applications, the aviation domain, compared to other domains, has still been relatively slow in adopting ML methods. Although we have abundant aviation data such as flight tracks, weather, safety reports, etc., the data are seldom in a ready-to-use format for the application of machine learning algorithms. In addition, those data are high-dimensional and heterogeneous. For example, a flight track may be comprised of thousands of time sequence points (position reports), and different tracks may have a different number of points. The current practice of manually processing features is labor-intensive, does not scale well to new problems, and is prone to information loss, affecting the effectiveness and maintainability of ML procedures. For that, this study proposed an unsupervised learning method, autoencoder, to automatically learn effective features from data.

In the aviation domain, flight-track data play a vital role for many studies. At MITRE, many studies have the desire to filter helicopter tracks out of collected flight track data for studying commercial flights or general aviation flights. The current method for helicopter identification mainly relies on the aircraft type or callsign matching, which faces a lot of challenges due to missing or mislabeled aircraft types in the data. To resolve the current perplexity, this research leveraged autoencoder to learn characteristics of helicopter tracks with partially available labeled data, and then apply it to identify helicopter tracks from unlabeled tracks. The results show that this novel method can successfully identify more helicopter tracks than the current rule-based method.

The remainder of the paper is organized as follows: Section II provides a brief introduction to autoencoders, Section III presents the helicopter identification autoencoder, data collection is described in Section IV, followed by a discussion of the results in Section V. Finally, concluding remarks are in Section VI.

## II. Autoencoder

An autoencoder is a neural network designed for the task of representation learning in an unsupervised way [6]. It tries to reconstruct the original input while compressing the data in the process to discover a more efficient and compact representation (Figure 1). The idea was originated in the 1980s, and later promoted by the seminal paper of Hinton and Salakhutdinov [7]. An autoencoder consists of two parts:

- **Encoder**: Compresses the original high-dimension input into the latent low-dimensional representation.
- **Decoder**: Attempts to reconstruct the data from the latent representation.

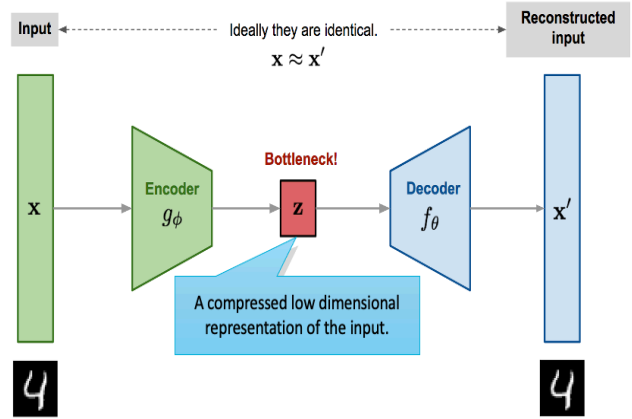

**Figure 1. Illustration of the process and architecture of the autoencoder model [6].**

By giving the input $X$ to the encoder, and with the decoder outputting $X'$, the neural network is trained to make $X$ and $X'$ nearly identical, with the difference measured by the reconstruction error. The bottleneck $Z$ is a compressed low dimensional representation of input $X$, which can then be used to support ML modeling. In our study, mean absolute error (MAE) is used to measure the reconstruction error (Eq. 1).

$$\text{MAE} = \sum_{i=1}^{n}(x_i - x'_i)/n \qquad (1)$$

## III. Helicopter Identification Detection Autoencoder

Flight-track data are very popular in the aviation domain for studies such as safety risk prediction, traffic flow management, and procedure-based navigation route design. At MITRE, many studies require filtering of helicopter tracks out of collected flight track data for studying commercial flights or general aviation flights. The current rule-based helicopter identification method heavily relies on the aircraft type or callsign matching, which is difficult because a lot of tracks have missing or mislabeled aircraft types in the data.

In an attempt to improve filtering accuracy, we chose an autoencoder to identify the helicopter tracks for its supremacy in representation learning from complex data. To build our helicopter track identification detection autoencoder, we have designed a four-step procedure:

**Step 1.** Feed preliminary "labeled" helicopter data to train autoencoder models.

In this step, the autoencoder will automatically learn the data representations of the helicopter tracks.

**Step 2.** Identify MAE reconstruction error threshold (Figure 2).

Previously, we introduced MAE as a measurement of difference between $X$ and $X'$. We plot the histogram of MAE and then choose the proper percentile (e.g., 80%) as an identification threshold. All tracks with MAE less than the threshold are considered to represent common helicopter behaviors. The others are considered as outliers. The proper threshold should minimize both false positive and false negative.

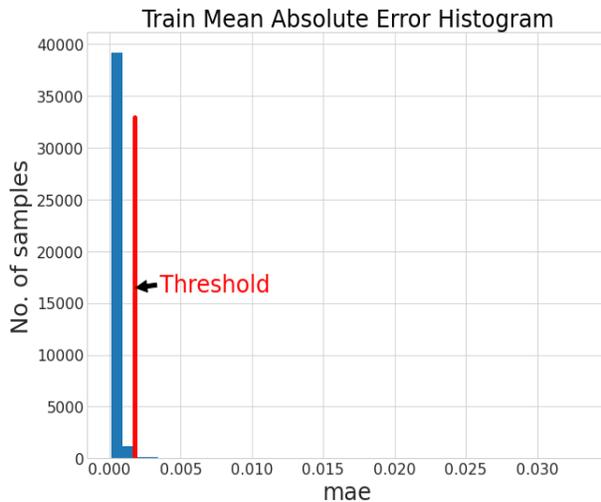

**Figure 2. Mean absolute error histogram of training dataset**

**Step 3.** Test model.

Run model for testing dataset and check the following two conditions listed in Figure 3 to decide whether a track is from a helicopter or not. The first condition is that the reconstruction error MAE is less than threshold δ defined in step 3; the second condition is that arrival runway score is smaller than the defined threshold Δ. The arrival runway score represents the confidence that the aircraft will land on the runway and is calculated while considering factors such as distance from runway threshold, runway length, course difference between candidate point and runway centerline course, lateral deviation from runway centerline, and if runway is reported in a scratchpad. The second condition is to remove general aviation (GA) flight tracks, which are not easy to distinguish from helicopter tracks. For that reason, we require the track must satisfy both of two conditions to qualify as a helicopter track. We should note that some helicopters could have large runway score due to imperfection of our runway score calculation algorithm; therefore, the second condition can also result in filtering some helicopters out.

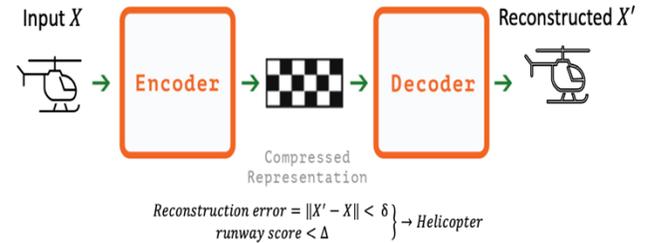

**Figure 3. Helicopter track identification autoencoder illustration**

**Step 4.** Validate the results.

To validate our identification results, we use an aircraft registration database, which can provide information such as aircraft tail number, mode-s code, aircraft type, and aircraft class (e.g., helicopter, fixed wing) ([8]). We joined track data with aircraft registration database on aircraft tail number or mode-s code and then use aircraft class column to confirm if the track is helicopter. Figure 4 shows an example of our validation results. The green dots are representing tracks which are validated as true, and the red x is representing those validated as false, and the orange square is ones which can't be validated from aircraft registration data. From the plot, you can see that our autoencoder did a good job on identifying the helicopter tracks. The plot can also help to adjust thresholds in step 3.

If necessary, repeat step 2 and 3 until the desired performance is achieved.

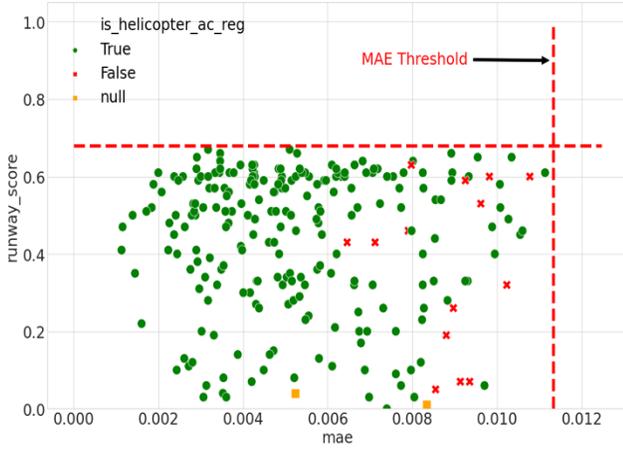

**Figure 4. Helicopter identification results validation**

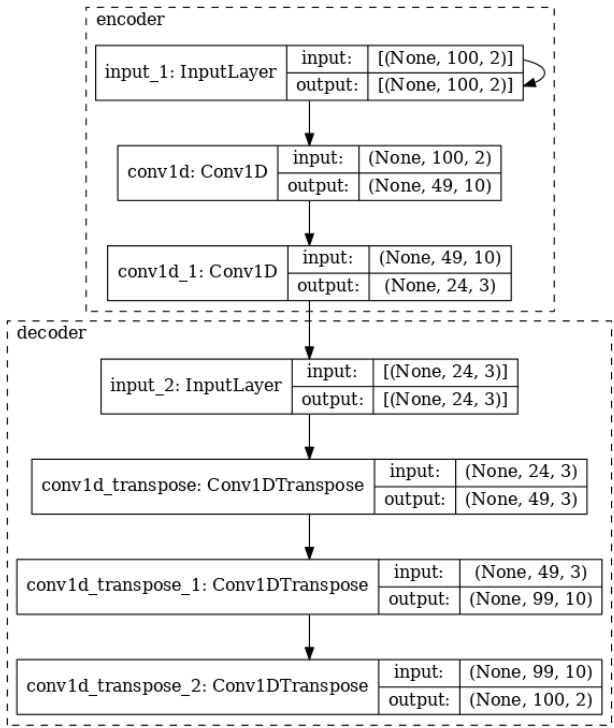

**Figure 5. Flight track helicopter detection autoencoder network structure**

With consideration that track data are long sequence time series, we built our autoencoder with a 1D convolution neural network (CNN) (e.g., CONV1D in TensorFlow [9]). We used keras and TensorFlow (version 2.4.0) to develop our autoencoder models [10]. For autoencoders, the shape of inputs and output should be same. Therefore, we selected CONV1D to build our encoder, and CONV1DTranspose to build our decoder. Figure 5 presents the autoencoder model architecture used in this effort.

## IV. Data Collection

**Table 1. Transportation Data Platform (TDP) Data Sources**

| Dataset | Function |
|---|---|
| National Flight Data Center (NFDC) | Airport and runways infrastructure data table with information such as:<br>• Airport latitude, longitude, elevation<br>• Runway threshold latitude, longitude, and elevation |
| Threaded Track | Data from several different surveillance sources are fused into a single "best" representation of a flight's track. The data table with information such as:<br>• Time, latitude, longitude, altitude, course, speed etc.<br>• Aircraft type |
| Airport Runway Assignment | Identifies probable landing runway based on track data points |
| Missed Approach | Identifies flights with missed approach |
| Aircraft Type | Aircraft type information such as helicopters, military ones, and unmanned aircraft system (UAS) derived from FAA's Order 7360 [12] |
| Aircraft Registration | Contains the records of all U.S. Civil Aircraft maintained by the FAA, Civil Aviation Registry, Aircraft Registration Branch, AFS-750 |

For our research, we identified five datasets in MITRE's Transportation Data Platform (TDP) [11], where a suite of surveillance data and supporting datasets including infrastructural and operational information are provided. Table 1 lists the five datasets and summarizes their functionalities.

**Error! Reference source not found.** shows the extract, transform, load (ETL) process used in this effort.

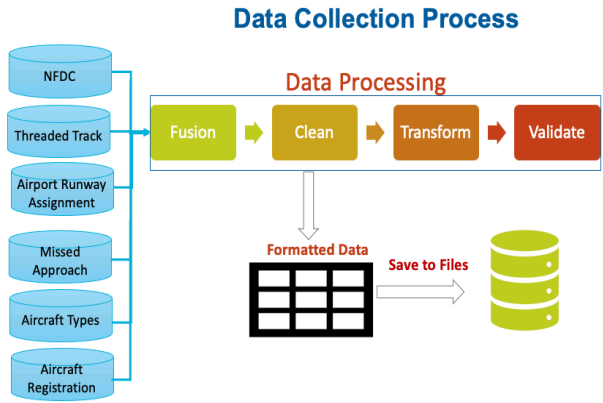

**Figure 6. Data collection process**

For the training process, we processed arrival track segments in terminal airspace, and then for each track, we collected the last 100 track points prior to the runway threshold as our training data.

## V. Results

In this section, we present helicopter identification results. We trained models using aircraft track data collected on various airports one at a time.

Table 2 lists seven airports we have trained models for. About 10% of all operations at McCarran International Airport (LAS) are helicopters. These helicopter operations have very common routes. Also, a large percentage of Grand Canyon West Airport (1G4) operations are helicopters. Therefore, training the models for these airports was easier. Dallas Love Field (DAL), Ronald Reagan Washington National (DCA), Van Nuys (VNY), Teterboro (TEB), and Phoenix Deer Valley (DVT) airports all have relatively small numbers of helicopter operations, which made training more difficult because of smaller data size. Figure 7 summarizes the training difficulties by airport.

Figure 8 shows an example of the helicopter identification result for a track. From that, we can see the current method didn't recognize it as a helicopter. With our autoencoder method, we can identify it as a helicopter. Moreover, the aircraft registration data also verifies that our result is correct. The aircraft registration data can be used for validation purposes because for some tracks we can find a match in this database and confirm aircraft class. However, there are many track instances without associated tail number or mode-s code, so those can't be matched; therefore, an alternate method is needed. In addition, the old aircraft type listed in the data is not valid. Through our validation process, we can find the correct aircraft type for this track, which is more useful to analysts or domain experts.

**Table 2. Airports for which trained models were developed**

| FAA Code | Airport Name |
|---|---|
| LAS | McCarran International Airport |
| DAL | Dallas Love Field |
| DCA | Ronald Reagan Washington National Airport |
| DVT | Phoenix Deer Valley Airport |
| TEB | Teterboro Airport |
| VNY | Van Nuys Airport |
| 1G4 | Grand Canyon West Airport |

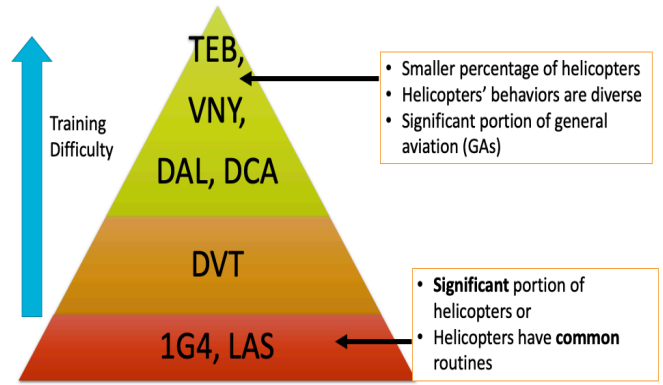

**Figure 7. Training difficult ordered by airports**

Figure 9 compares helicopter identification results for 1G4 airport in 06/2019. In the graph, the small brown circle shows that current rule-based method identifies 70 (67+3) helicopters, and the big yellow circle indicates that autoencoder finds 959 (892+67) helicopters. Sixty-seven helicopters are identified by the two methods. However, there are 3 helicopters which current method has identified, but the autoencoder missed. In contrast, 892 helicopters are identified by autoencoder alone. Therefore, the autoencoder identifies thirteen times as many helicopters as the current method.

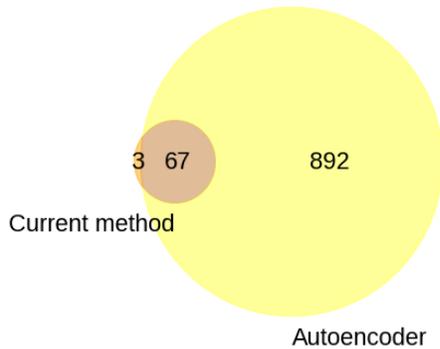

**Figure 8. An example of helicopter identification results**

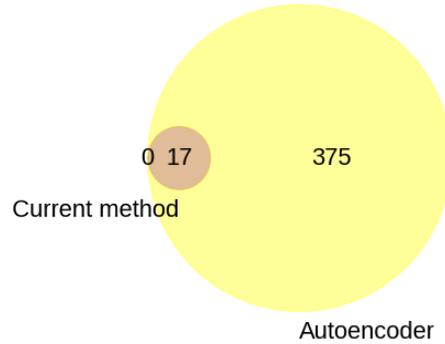

Figure 9. 1G4 helicopter identification results comparison in 06/2019

Similarly, Figure 10 compares helicopter identification results for DVT airport in 10/2019. At DVT, about 70% of flight tracks have missing aircraft types in the track data. From the graph, you can see that current rule-based method just can identify 17 helicopters. However, our autoencoder method can identify 392 (375+17) helicopters which also have also been validated as accurate. Therefore, in this case, autoencoder can identify 22 times more helicopter operations than the current method.

**Figure 10. DVT helicopter identification results comparison in 10/2019**

We should point out that the autoencoder does not work as well at the other 5 airports tested, as they have traffic mixes that are more difficult to train on. For example, VNY and TEB have a lot of general aviation flights, which are often difficult to distinguish from helicopters. The helicopters at DAL and DCA airports have very diverse behaviors, which makes it harder for autoencoder to learn an effective data representation from this smaller dataset.

Another benefit of our study is that our validation process through aircraft registration dataset can also find meaningful aircraft types for track with pseudo helicopter types such as HELO or HELI in the data. Figure 11 shows an example of 5 new aircraft types we have identified at 1G4, and Figure 12 presents an example of 13 new aircraft types we have identified at TEB. These new types can be used to support of our current helicopter identification method.

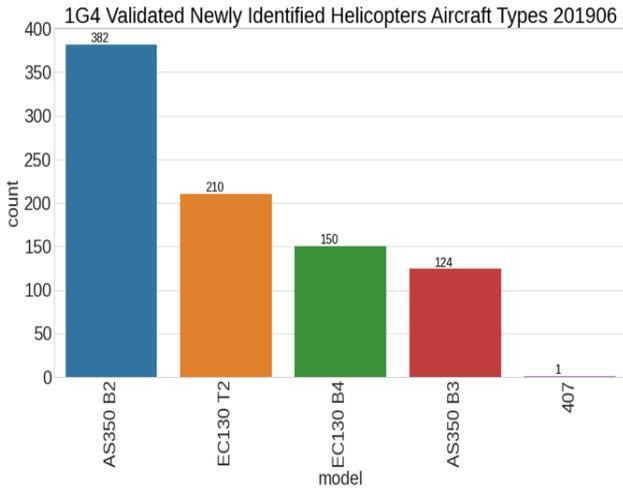

**Figure 11. Newly identified aircraft types for flight tracks at 1G4**

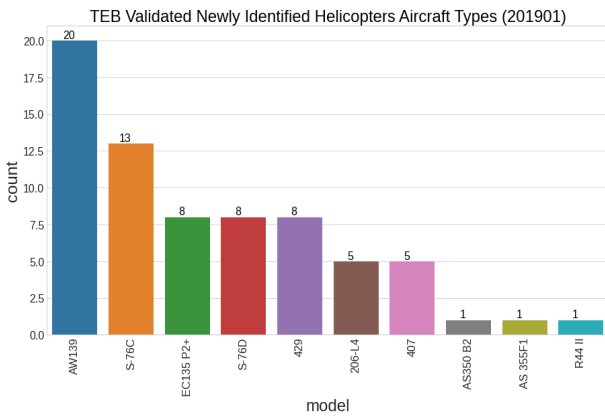

**Figure 12. Newly identified aircraft types for flight tracks at TEB**

## VI. Conclusion

This paper employed a novel unsupervised learning method, an autoencoder, for helicopter identification, exploiting the autoencoder's powerful data representation learning capability. We trained models for several airports with widely varying traffic characteristics. The results show that an autoencoder could successfully learn effective representation for flight track data across these varying conditions. For example, at DVT airport, where 70% of tracks have missing aircraft types, the autoencoder was able to identify twenty-two times more helicopters than the current method; for 1G4 airport, the autoencoder can identify thirteen times more helicopters than the current approach. Our study can also identify mislabeled aircraft types in the track data and find true types for records with pseudo aircraft type labels such as HELO or HELI. With improved labelling, studies using these data sets can produce more reliable results. We also summarized the causes of modeling difficulty at different airports.

## Acknowledgments

We thank the following MITRE colleagues: Clark Lunsford, Dr. Alex Tien, Mike Robinson, Dan Larson, Joe Hoffman, Karl Mayer, Van Hare, Eric Zakrzewski, Adric Eckstein, Gaurish Anand, Evan McClain, Paul Diffenderfer, and Dr. Lakshmi Vempati for the valuable discussions and insights.

## NOTICE

## References

[1] Bengio, Yoshua, Aaron Courville and Pascal Vincent, 2013, Representation Learning: A Review and New Perspectives, vol. 35, no. 8, IEEE Transactions on Pattern Analysis and Machine Intelligence, pp. 1798-1828.

[2] Smith, Andrew, Kyle Collins and Dimitri Mavris, 2017, Survey of Technology

Forecasting Techniques for Complex Systems, Grapevine, Texas, 58th AIAA/ASCE/AHS/ASC Structures, Structural Dynamics, and Materials Conference.

[3] Wang, Zhenming, Sherry Lance and John Shortle, 2016, Improving the Nowcast of Unstable Approaches, Dullas,VA, Integrated Communications, Navigation, Surveillance (ICNS).

[4] Maheshwari, Apoorv, Navindran Davendralingam and Daniel DeLaurentis, 2018, A Comparative Study of Machine Learning Techniques forAviation Applications, Altalanta, GA, Aviation Technology, Integration, and Operations Conference.

[5] Kim, Young Jin, Choi Sun, Briceno Simon, and Mavris Dimitri, 2016, A deep learning approach to flight delay prediction, Sacramento, CA, IEEE/AIAA 35th Digital Avionics Systems Conference (DASC).

[6] Weng Lilian, 12 Aug 2018, From Autoencoder to Beta-VAE, [Online], Available: https://lilianweng.github.io/lil-log/2018/08/12/from-autoencoder-to-beta-vae.html. [Accessed 10 Oct 2020].

[7] Hinton, Geoffrey and Ruslan Salakhutdinov, 2006, Reducing the Dimensionality of Data with Neural Networks, vol. 313, no. 5786, Science, pp. 504-507.

[8] Federal Aviation Administration, Federal Aviation Administration Registration, [Online]. Available: https://registry.faa.gov/aircraftinquiry/Search/NNumberInquiry. [Accessed 10 Aug 2020].

[9] Google, Tensorflow Conv1d, [Online], Available: https://www.tensorflow.org/api_docs/python/tf/keras/layers/Conv1D . [Accessed 22 Oct 2020].

[10] Google, Tensorflow, [Online], Available: https://www.tensorflow.org/. [Accessed 10 Aug 2020].

[11] Eckstein, Adric, Chris Kurcz and Marcio Silva, 2012, Threaded Track: Geospatial Data Fusion for Aircraft Flight Trajectories, McLean, VA, The MITRE Corporation.

[12] Federal Aviation Administration, Federal Aviation Administration order 7360, [Online], Available: https://www.faa.gov/regulations_policies/orders_notices/index.cfm/go/document.information/documentid/1030848. [Accessed 22 Oct 2020].